# Reliability Assessment of Neural Networks in GPUs: A Framework For Permanent Faults Injections

Juan-David Guerrero-Balaguera*, Luigi Galasso*, Robert Limas Sierra+, Matteo Sonza Reorda*
*Politecnico di Torino, Department of Control and Computer Engineering (DAUIN)
{juan.guerrero, luigi.galasso, matteo.sonzareorda}@polito.it
+Universidad Pedagógica y Tecnologica de Colombia (UPTC), Electronics Engineering Scholl
{robert.limas}@uptc.edu.co

*Abstract* — Currently, Deep learning and especially Convolutional Neural Networks (CNNs) have become a fundamental computational approach applied in a wide range of domains, including some safety-critical applications (e.g., automotive, robotics, and healthcare equipment). Therefore, the reliability evaluation of those computational systems is mandatory. The reliability evaluation of CNNs is performed by fault injection campaigns at different levels of abstraction, from the application level down to the hardware level. Many works have focused their effort on evaluating the reliability of neural networks in the presence of transient faults. However, the effects of permanent faults have been investigated at the application level, only, e.g., targeting the parameters of the network. This paper intends to propose a framework, resorting to a binary instrumentation tool to perform fault injection campaigns, targeting different components inside the GPU, such as the register files and the functional units. This environment allows assessing the reliability of CNNs deployed on a GPU considering the presence of permanent faults.

## I. INTRODUCTION

Nowadays, Artificial Neural Networks (ANNs) are bioinspired computational models widely deployed in multiple application domains, such as multimedia, financial analysis, robotics, and automotive [1]. Convolutional Neural Networks (CNNs) represent a class of ANNs used to implement complex image and video processing algorithms for object recognition and path tracking [2]. These CNN-based algorithms are widely used in safety-critical systems, such as self-driving vehicles [1], [2]. Therefore, estimating the reliability of those systems is crucial to meet the requirements of different safety standards (e.g., ISO26262 for the automotive domain).

CNNs models are implemented on a wide range of hardware accelerators [3], including Graphics Processing Units (GPUs), Tensor Processing Units (TPUs) [4], FPGAs [5], and ASICs. However, GPUs are the dominating devices among these accelerators and are extensively adopted to implement and accelerate CNNs. GPUs are the preferred accelerator in many applications, mainly due to their architecture, computational power, flexibility, and scalability. Also, some GPU devices include custom hardware modules specially designed to accelerate cutting-edge CNNs architectures (e.g., tensor cores).

In other cases, the GPUs are incorporated in specialized hardware and software ecosystems, including Deep Learning development environments, aiming for safety-critical applications such as autonomous vehicles [6], which demand high-reliability standards keeping low failures rates as much as possible. However, the cutting-edge semiconductor technologies used for GPUs' manufacturing are susceptible to suffer faults, including permanent faults arising in the device during its operational life (e.g., due to aging) [7]. Thus, faults in the GPU device may affect a CNN model generating wrong results or catastrophic events at the application level.

Historically, the ANNs have been considered as intrinsically fault-tolerant due to the redundancy in their structure and connections, allowing them to tolerate some noise generated by the input data or during the computational process [8]. However, modern CNNs have high topological complexity, including a deeper hierarchy of layers and more operations than classical ANN architectures. Thus, the traditional methods based on error propagation analysis cannot be applied to new state-of-the-art CNNs. Instead, new studies about the reliability of CNNs consider that the fault tolerance of those computational models depends on the hardware that executes the CNN and the topology of the network [8]. Therefore, the reliability analysis of CNNs implemented on GPUs deserves special attention, considering that both CNNs and GPUs are popularly used to implement many safety-critical applications; thus, any fault presence can jeopardize them.

According to the literature, the resilient evaluation of neural networks to faults is usually carried out by performing Fault Injection Campaigns (FICs) at different abstraction levels, from the application level down to the hardware level.

So far, many works [9]–[11] have focused their effort on evaluating the reliability of neural networks in the presence of transient faults models. However, permanent fault models have been addressed using application-level FICs, only targeting the parameters of the network (weights, activation values, etc.) [12]. Unfortunately, this fault injection approach differs from reality by far at this level. Since this approach allows only injecting faults in the system memory, generating a significant limitation when considering fault injection in other hardware components such as register files, functional units, and control units.

On the other hand, hardware-level FIC resorts to injecting faults in the circuit at the RTL or gate-level. However, the lower the abstraction level, the higher the fault injection time required [13]. This FIC approach is more realistic but imposes serious limitations in terms of injection time due to the circuit size (millions of gates for GPUs) plus the network architecture (3M of weights and 9K of neurons for small CNN LeNet-5). Therefore, the reliability analysis during the inference of a CNN at the hardware level clearly requires an excessive amount of time [13]] and computational resources. Consequently, an alternate solution to be considered relies on the architectural-level fault injections using Software Implemented Fault Injections (SWIFI) [14], also known as Hardware Injection Through Program Transformation (HIPT) [15] for GPUs contexts. These approaches consist of modifying the application's source code as ISA level to inject errors, and then the program is issued in the real hardware device.

To the best of our knowledge so far, there are no reported works in the literature considering frameworks HIPT-based for FICs of *permanent faults* oriented to evaluate the reliability of CNNs on GPUs. Therefore, this paper intends to propose a fault injection framework, resorting to implementing FIC of permanent faults on GPU devices by injecting errors at the architectural level. The environment allows injecting errors in the register files and the functional units (integers and floating-point units). In particular, the proposed framework was built using the NVIDIA NVBITFI tool. This tool was adapted and customized, extending its capabilities to support the injection and propagation of hard errors that mimic the presence of permanent faults in a specific hardware component of the GPU.

The rest of the paper is organized as follows. Section II introduces the essential background and related works in the field. Section III presents the proposed framework, and Section IV describes the study cases and the experimental setup. Section V provides the experimental results. Finally, Section VI draws some conclusions and future research directions.

## II. BACKGROUND

### A. Convolutional Neural Networks

A Convolutional Neural Network (ConvNet/CNN) is one class of ANNs mainly applied to video and image processing to pattern recognition and image classification. The architecture of a CNN mimics the connection patterns of the neurons in the visual cortex of the Human Brain. A CNN consists of an input layer, hidden layers, and output layer. Unlike the multilayer perceptron (MLP), the hidden layers in a CNN perform convolutional operations between the convolution kernel (weights) and the input matrix. Each layer also incorporates activation functions to bound the output values of the neurons. Additionally, each convolutional layer is followed by other layers such as pooling, fully connected, and normalization layers [8]. The well-known CNN architectures involve LeNet, AlexNet, VGGNet, GoogLeNet, ResNet, and DenseNet.

### B. GPU organization

Modern Graphics Processing Units (GPUs) are devices composed of scalable arrays of parallel execution units known as Streaming Multiprocessors (SMs). Each SM in a GPU uses the Single-Instruction Multiple-Tread (SIMT) execution model, where a single instruction executes multiple independent threads concurrently.

A scheduler unit controls the execution of one SIMT instruction, fetching, decoding, and distributing the workloads to be executed in the parallel processing cores called Streaming Processors (SPs) of the SM. One SM can contain between 8 and 128 SPs depending on the GPU model and the number of parallel threads to be processed concurrently. Additionally, Each SM includes one or more scheduler controllers to manage and trace the assigned tasks. The parallel task is divided into threads and groups of threads (32 Threads) (called Warps) and submitted for execution on each SM. Internally, the SM also includes local memories and register file banks to process each thread in parallel. In modern SM architectures, the number of available execution modules is proportional to the granularity of a warp.

### C. Hardware Injection Through Program Transformation (HIPT)

The HIPT architectural-level fault injector tools mimic the hardware fault effect as an error by implementing instrumentation functions in the target application's source code. These errors are inserted, corrupting the output result of an executed instruction on a selected core in the GPU. Once instrumented the source code, the faulty application is executed in the device, propagating the error at hardware speeds to the output of the target application [15], [16]

NVBitFI is the state-of-the-art fault injection tool that instruments a target program to inject errors into NVIDIA GPUs. This tool performs the instrumentation at the SASS (Stream Assembly). Unlike other tools such as SASSIFI, NVBitFI allows to instrument unknown libraries during the build time. NVBitFI offers a single interface to support the recent NVIDIA architecture families, including Kepler, Maxwell, Pascal, Volta, and Ampere GPUs [15]. NVBitFI incorporates four bit-flip models for transient faults injection. 1) single bit-flip, 2) two-adjacent bit-flip, 3) random value, 4) all zero. Additionally, this tool incorporates a simplistic approach for injecting permanent faults in the functional units, applying a random bit-flip to the result of all instances of one selected instruction opcode. Therefore, in this work, we propose to extend the existent fault injection capabilities of the tool to support permanent fault injection in CNN models.

## III. PROPOSED FRAMEWORK FOR PERMANENT FAULTS INJECTION

The proposed environment uses the NVBitFI instrumentation tool to perform permanent fault injection campaigns in a GPU executing a CNN model. The framework uses a global controller to manage the fault injection process. A pre-trained CNN model is used during the fault injection process performing the inference of a subset of images that are not used during the training stages. The fault injection of permanent faults is carried out as follows. First, the controller initiates the inference of the CNN model under a fault-free scenario in order to collect its outputs and create a golden reference model. Then, one fault is injected at a time, propagating it through all executed kernels in the GPU. Finally, the results are collected and compared with the golden model outcomes to decide the classification of the injected error in the CNN algorithm.

The permanent faults can produce different effects on the outputs of the neural network. Those effects can be classified according to their impact in the CNN. Some propagated errors can be considered Silent Data Corruption (SDCs) because they are propagated to the output modifying the inference results. This type o error sometimes is critical since the expected classification changes at all. Other errors produce a critical failure during the execution of the application causing a Detected Unrecoverable Error (DEU). This type of error suddenly stops the CNN computations, and the inference process does not finish. Finally, an output error is considered Masked when it does not produce any output effect or is mitigated during the inference process. In this work, we considered four error categories as follows:

**SDC- Critical**: The injected error propagates to the output of the CNN, modifying the probabilities vector during the inference calculation producing a misclassification result.

**SDC-Safe**: The injected error propagates to the output of the CNN modifying the probabilities vector during the inference calculation, but the classification output is still correct.

**DUE**: The injected error produces a system hang or crash. This error interrupts the execution of the CNN at any time. The causes of this behavior can be memory access violation, memory misalignment violation, or timeout (the error block the CNN model in an infinite loop).

**Masked**: the injected error does not have any impact on the output. In this case, the output is the same as the fault-free scenario

The fault injection campaigns at the GPU level are issued through the binary instrumentation tool mechanism by NVBitFI. This tool is designed to intercept and instrument, with errors, the launched kernels of the target application. Three steps are used to perform this instrumentation process. i) instructions inspection, ii) insertion of instrumentation function, and ii) kernel execution. In the inspection instruction step, each SASS instruction of the kernel is examined in order to extract detailed information such as the instruction number, opcode type, registers identifiers, and memory references. In the second step, the extracted information of the instruction is compared with the injection specifications given by the user according to the selected error model (opcode, instruction number, destination registers, etc.). Therefore, if the instruction matches the target specifications, an error injector function is inserted immediately after the matched instruction. The injector function writes in the associated destination register the desired error by inserting bit-flips or modifying the instrumented instruction result entirely. When this process finishes, the instrumented kernel is executed in the device to propagate the injected error(s) to the next kernel. This mechanism is repeated for every kernel of the application.

The permanent faults modeling using the NVBitFI tool can be performed assuming that the presence of the fault in a particular module of the GPU affects all instructions in the application sharing the faulty module. Thus, this modeling process requires different strategies for each hardware component. In this work, we integrate two additional fault injectors to NVBitFI. The first injector is specially designed to model permanent faults in the GPU register files. The second one targets the functional units using a realistic approach that the one considered initially by NVBitFI.

*A. Permanent Fault injection in the Register Files*

During the parallel execution of an application on a GPU, the block scheduler distributes on each SM several blocks of threads to be executed in parallel. This distribution follows the maximum occupancy of the device to maximize the GPU's performance. Then, each block of threads is distributed in the available warps. The GPU architecture defines the maximum number of warps residents per SM simultaneously. Additionally, each resident thread of each resident warp has access to a private set of registers to perform the needed operations of the application.

Therefore, to model a permanent fault in the registers of the GPU, it is required to specify not only the instruction and the register number, but also the warp identifier and the thread inside the selected warp to maintain the fault effect into the same location during the whole application execution. Consequently, the permanent fault injection in any register of the GPU can be modeled using the exact mechanism of NVBitFI described before, targeting for instrumentation only those instructions using the faulty target register as the destination.

The fault definition is composed of the quintuple <SMID, threadID, Register, Mask, stuck-at> where SMID represents the SM where the injection should be performed out of the many possibly available; threadID is the resident thread selected in which to realize the injection; Register is the target destination register to be injected; Mask is the single bitmask to be applied to the target register value; stuck-at can be 1 or 0 depending on the value to be forced in the defined bit.

*B. Permanent Fault injection in the Functional Units*

Similarly, the fault injection of permanent fault in the functional units follows the same steps as the register files. However, there is a slight difference when we consider a permanent fault on an arithmetic unit. Those faults might affect the results of several instructions opcodes sharing the same operational unit. For example, suppose the floating-point unit is considered affected by a permanent fault. In that case, the instrumentation process shall consider the injection of errors (bit-flips) in the output of each floating-point instruction issued by each kernel of the CNN.

## IV. EXPERIMENTAL RESULTS

The proposed fault simulation framework was evaluated using the pre-trained LeNet and AlexNet CNNs models available in the Darknet environment [17]. The trained LeNet model can classify ten handwritten digits (0 to 9) using the MNIST dataset. The AlexNet model classifies images from 1000 categories using the ImageNet dataset. A set of fault simulation experiments was performed to evaluate the impact of permanent fault on the selected CNNs. The fault simulation campaigns were performed on a workstation HP Z2 G5 with CPU Intel Core i9-10800 20 cores, 32 GB of RAM memory, and equipped with an RTX 3060TI GPU platform including an NVIDIA Ampere architecture with compute capability (CC) 8.6. On the other hand, the LeNet model was evaluated using the embedded platform Jetson Nano, which has an NVIDIA Maxwell architecture and CC 5.3.

For the experiments, the LeNet model was evaluated by performing fault injections on the register file and the functional units on one SM. In the case of the AlexNet model, the fault injections were performed considering permanent faults in the register file only. During the fault simulation campaigns, 29000 faults were considered for the functional units using a single-bit flip scenario, and 16000 for the register files considering only the first ten registers (R0 to R9). The register selection was made based on the analysis results using the profiler tool present in NVBitFI, which allows us to calculate the frequency of usage of each register during the execution of the CNNs models.

TABLE I. RELATIVE OCCURRENCE OF ERROR FOR PERMANENT FAULTS IN THE REGISTER FILE OF A GPU

| CNN Model | GPU arch | DEU (%) | SDC safe (%) | SDC Critical (%) | Masked (%) |
|---|---|---|---|---|---|
| LeNet | Maxwell | 74.76 | 10.01 | 2.73 | 12.48 |
|  | Ampere | 64.07 | 16.51 | 0.51 | 18.90 |
| AlexNet | Ampere | 63.36 | 15.26 | 2.70 | 18.68 |

TABLE II. RELATIVE OCCURRENCE OF ERROR FOR PERMANENT FAULTS IN THE FUNCTIONAL UNITS

| CNN Model | Funct unit | DEU (%) | SDC safe (%) | SDC Critical (%) | Masked (%) |
|---|---|---|---|---|---|
| LeNet | INT core | 24.82 | 6.38 | 2.21 | 66.58 |
|  | FP core | 3.96 | 3.92 | 0.56 | 91.56 |
|  | SFU cores | 16.93 | 2.98 | 10.37 | 69.71 |

Tables I and II report the results of the fault injection campaigns performed for the register files and functional units targeting one SM of the GPU. Both tables report the occurrence of errors according to the categories mentioned before DEUs, SDCs, and Masked. Moreover, the tables present the result for different GPUs in the case of the register file.

The reported results in table I show that permanent faults in the registers can produce a significant number of DUEs occurrences exceeding 60% of the cases. Even for the LeNet CNN, the percentage of occurrence reached 74.76% when a Jetson Nano GPU was considered. Interestingly, regardless of the CNN evaluated, the occurrences of DEUs are lower by approximately 10% when an Ampere GPU is used instead of the Maxwell one. Similarly, SDC safe and masked occurrences are around 6% lower on an Ampere GPU than in a Maxwell. The SDC critical for LeNet differs drastically from 2.73% considering Maxwell GPU to 0.51% on Ampere.

These results linked to the GPU architecture can be explained considering the hardware details of each GPU used in the experiments. In the case of Maxwell architecture, on the Jetson Nano board, the embedded GPU only disposes of one SM to execute all the kernel blocks issued by the CNN application. Therefore, one permanent fault in the registers has more probability of corrupting many threads, increasing the occurrences of DEUs and SDCs. On the other hand, the Ampere GPU has 38 SMs. Therefore, one permanent fault affecting only one SM generates propagation of the results in a few cases. It's worth noticing that the DEUs in one SM affect the complete application regardless of the number of SMs available in the GPU device.

Table II shows that permanent faults in the integer cores and SFU cores produce up to 24% of DEU occurrences. Only the Floating-point unit has a low DUE rate occurrence, less than 4%. It is worth noticing that the SDC safe occurrence is lower than 6% for all the evaluated units. Interestingly, the SDC critical reaches around 10% for the SFU cores. Surprisingly, the occurrence rate of Masked faults is greater than 60% for all the cases. The reason for those results in the functional units is due to the fact that we only use the single-bit flip injection; therefore, the results show a more optimistic scenario than the register file experiments.

V. CONCLUSIONS

In this work, we extend the capabilities of the NVBitFI tool, developed by NVIDIA, to inject permanent faults in order to support the reliability evaluation of CNNs implemented on GPU devices. For the experiments, two pre-trained CNN models were considered LeNet and AlexNet. The fault injection campaigns considered permanent faults in the register file of one SM of the GPU and permanent faults in the functional units. The results show that permanent faults on a GPU produce a high rate of critical failures, generating a high rate of occurrence of DUEs up to 74.76%. Similarly, for the functional units, the DUE rate reaches 24%. Regarding misclassification results due to permanent faults, in the case of the register file, the results show a maximum of 2.73%; however, for SFU core, these results increase up to 10.30%. Therefore, as stated before, when considering permanent faults at lower levels, closer to the actual device, the propagated effect of those faults to the outputs shows a more realistic scenario considering occurrences of critical errors such as DUEs and SDC critical.

Future activities aim to evaluate more CNN architecture models and propose hardening techniques to counteract the vulnerabilities caused by permanent faults in CNN architectures using GPUs.